\documentclass{article}
\usepackage{ijcai20}

\usepackage{times}
\usepackage{soul}
\usepackage{url}
\usepackage{hyperref}
\usepackage[utf8]{inputenc}
\usepackage[small]{caption}
\usepackage{graphicx}
\usepackage{amsmath}
\usepackage{amssymb}
\usepackage{amsthm}
\usepackage{booktabs}
\usepackage{algorithm}
\usepackage{algorithmic}
\hypersetup{hidelinks}

\title{Deeper Task-Specificity Improves Joint Entity and Relation Extraction}

\author{
    Phil Crone
    \affiliations
    Ancestry.com
    \emails
    pcrone@ancestry.com
}

\begin{document}

\maketitle

\begin{abstract}
    Multi-task learning (MTL) is an effective method for learning related tasks, but designing MTL models necessitates deciding which and how many parameters should be task-specific, as opposed to shared between tasks. We investigate this issue for the problem of jointly learning named entity recognition (NER) and relation extraction (RE) and propose a novel neural architecture that allows for deeper task-specificity than does prior work. In particular, we introduce additional task-specific bidirectional RNN layers for both the NER and RE tasks and tune the number of shared and task-specific layers separately for different datasets. We achieve state-of-the-art (SOTA) results for both tasks on the ADE dataset; on the CoNLL04 dataset, we achieve SOTA results on the NER task and competitive results on the RE task while using an order of magnitude fewer trainable parameters than the current SOTA architecture. An ablation study confirms the importance of the additional task-specific layers for achieving these results. Our work suggests that previous solutions to joint NER and RE undervalue task-specificity and demonstrates the importance of correctly balancing the number of shared and task-specific parameters for MTL approaches in general. 
\end{abstract}

\section{Introduction}

Multi-task learning (MTL) refers to machine learning approaches in which information and representations are shared to solve multiple, related tasks. Relative to single-task learning approaches, MTL often shows improved performance on some or all sub-tasks and can be more computationally efficient \cite{caruana1997multitask,Cipolla_2018,vandenhende2019branched,li2019empirical}. We focus here on a form of MTL known as {\em hard parameter sharing}. Hard parameter sharing refers to the use of deep learning models in which inputs to models first pass through a number of shared layers. The hidden representations produced by these shared layers are then fed as inputs to a number of task-specific layers. 


\begin{figure}[t]
    In 1809, author \textbf{Edgar Allan Poe}\textsubscript{\sc peop} was born in \textbf{Boston}\textsubscript{\sc loc}
        \caption{Example of a sentence containing named entities and relations from the CoNLL04 dataset. This sentence expresses a {\em Lives-In} relation between {\em Edgar Allan Poe} and {\em Boston}.}
    \label{fig:conll_ex}
\end{figure}

Within the domain of natural language processing (NLP), MTL approaches have been applied to a wide range of problems \cite{li2019empirical}. In recent years, one particularly fruitful application of MTL to NLP has been joint solving of named entity recognition (NER) and relation extraction (RE), two important information extraction tasks with applications in search, question answering, and knowledge base construction \cite{jiang2012information}. NER consists in the identification of spans of text as corresponding to named entities and the classification of each span's entity type. RE consists in the identification of all triples \((e_i, e_j, r)\), where \(e_i\) and \(e_j\) are named entities and \(r\) is a relation that holds between \(e_i\) and \(e_j\) according to the text. For example, in Figure \ref{fig:conll_ex}, {\em Edgar Allan Poe} and {\em Boston} are named entities of the types {\em People} and {\em Location}, respectively. In addition, the text indicates that the {\em Lives-In} relation obtains between {\em Edgar Allan Poe} and {\em Boston}.

One option for solving these two problems is a pipeline approach using two independent models, each designed to solve a single task, with the output of the NER model serving as an input to the RE model. However, MTL approaches offer a number of advantages over the pipeline approach. First, the pipeline approach is more susceptible to error prorogation wherein prediction errors from the NER model enter the RE model as inputs that the latter model cannot correct. Second, the pipeline approach only allows solutions to the NER task to inform the RE task, but not vice versa. In contrast, the joint approach allows for solutions to either task to inform the other. For example, learning that there is a {\em Lives-In} relation between {\em Edgar Allan Poe} and {\em Boston} can be useful for determining the types of these entities. Finally, the joint approach can be computationally more efficient than the pipeline approach. As mentioned above, MTL approaches are generally more efficient than single-task learning alternatives. This is due to the fact that solutions to related tasks often rely on similar information, which in an MTL setting only needs to be represented in one model in order to solve all tasks. For example, the fact that {\em Edgar Allan Poe} is followed by {\em was born} can help a model determine both that {\em Edgar Allan Poe} is an instance of a {\em People} entity and that the sentence expresses a {\em Lives-In} relation.

While the choice as to which and how many layers to share between tasks is known to be an important factor relevant to the performance of MTL models \cite{Zhao_2018,vandenhende2019branched}, this issue has received relatively little attention within the context of joint NER and RE. As we show below in Section 2, prior proposals for jointly solving NER and RE have typically made use of very few task-specific parameters or have mostly used task-specific parameters only for the RE task. We seek to correct for this oversight by proposing a novel neural architecture for joint NER and RE. In particular, we make the following contributions:
\begin{enumerate}
    \item We allow for deeper task-specificity than does previous work via the use of additional task-specific bidirectional recurrent neural networks (BiRNNs) for both tasks.
    \item Because the relatedness between the NER and RE tasks is not constant across all textual domains, we take the number of shared and task-specific layers to be an explicit hyperparameter of the model that can be tuned separately for different datasets.
\end{enumerate}

We evaluate the proposed architecture on two publicly available datasets: the Adverse Drug Events (ADE) dataset \cite{gurulingappa2012development} and the CoNLL04 dataset \cite{roth2004linear}. We show that our architecture is able to outperform the current state-of-the-art (SOTA) results on both the NER and RE tasks in the case of ADE. In the case of CoNLL04, our proposed architecture achieves SOTA performance on the NER task and achieves near SOTA performance on the RE task. On both datasets, our results are SOTA when averaging performance across both tasks. Moreover, we achieve these results using an order of magnitude fewer trainable parameters than the current SOTA architecture.

\section{Related Work}

We focus in this section on previous deep learning approaches to solving the tasks of NER and RE, as this work is most directly comparable to our proposal. Most work on joint NER and RE has adopted a BIO or BILOU scheme for the NER task, where each token is labeled to indicate whether it is the (B)eginning of an entity, (I)nside an entity, or (O)utside an entity. The BILOU scheme extends these labels to indicate if a token is the (L)ast token of an entity or is a (U)nit, i.e. the only token within an entity span.

Several approaches treat the NER and RE tasks as if they were a single task. For example, Gupta {\em et al.}~\shortcite{gupta-etal-2016-table}, following Miwa and Sasaki~\shortcite{miwa-sasaki-2014-modeling}, treat the two tasks as a table-filling problem where each cell in the table corresponds to a pair of tokens \((t_i, t_j)\) in the input text. For the diagonal of the table, the cell label is the BILOU tag for \(t_i\). All other cells are labeled with the relation \(r\), if it exists, such that \((e_i, e_j, r)\), where \(e_i\) is the entity whose span's final token is \(t_i\), is in the set of true relations. A BiRNN is trained to fill the cells of the table. Zheng {\em et al.}~\shortcite{Zheng_2017} introduce a BILOU tagging scheme that incorporates relation information into the tags, allowing them to treat both tasks as if they were a single NER task. A series of two bidirectional LSTM (BiLSTM) layers and a final softmax layer are used to produce output tags. Li {\em et al.}~\shortcite{li2019entity} solve both tasks as a form of multi-turn question answering in which the input text is queried with question templates first to detect entities and then, given the detected entities, to detect any relations between these entities. Li {\em et al.} use BERT \cite{devlin-etal-2019-bert} as the backbone of their question-answering model and produce answers by tagging the input text with BILOU tags to identify the span corresponding to the answer(s).

The above approaches allow for very little task-specificity, since both the NER task and the RE task are coerced into a single task. Other approaches incorporate greater task-specificity in one of two ways. First, several models share the majority of model parameters between the NER and RE tasks, but also have separate scoring and/or output layers used to produce separate outputs for each task. For example, Katiyar and Cardie~\shortcite{katiyar-cardie-2017-going} and Bekoulis {\em et al.}~\shortcite{bekoulis2018joint} propose models in which token representations first pass through one or more shared BiLSTM layers. Katiyar and Cardie use a softmax layer to tag tokens with BILOU tags to solve the NER task and use an attention layer to detect relations between each pair of entities. Bekoulis {\em et al.}, following Lample {\em et al.}~\shortcite{Lample_2016}, use a conditional random field (CRF) layer to produce BIO tags for the NER task. The output from the shared BiLSTM layer for every pair of tokens is passed through relation scoring and sigmoid layers to predict relations. 

\begin{figure*}
    \centering
    \includegraphics[width=\textwidth]{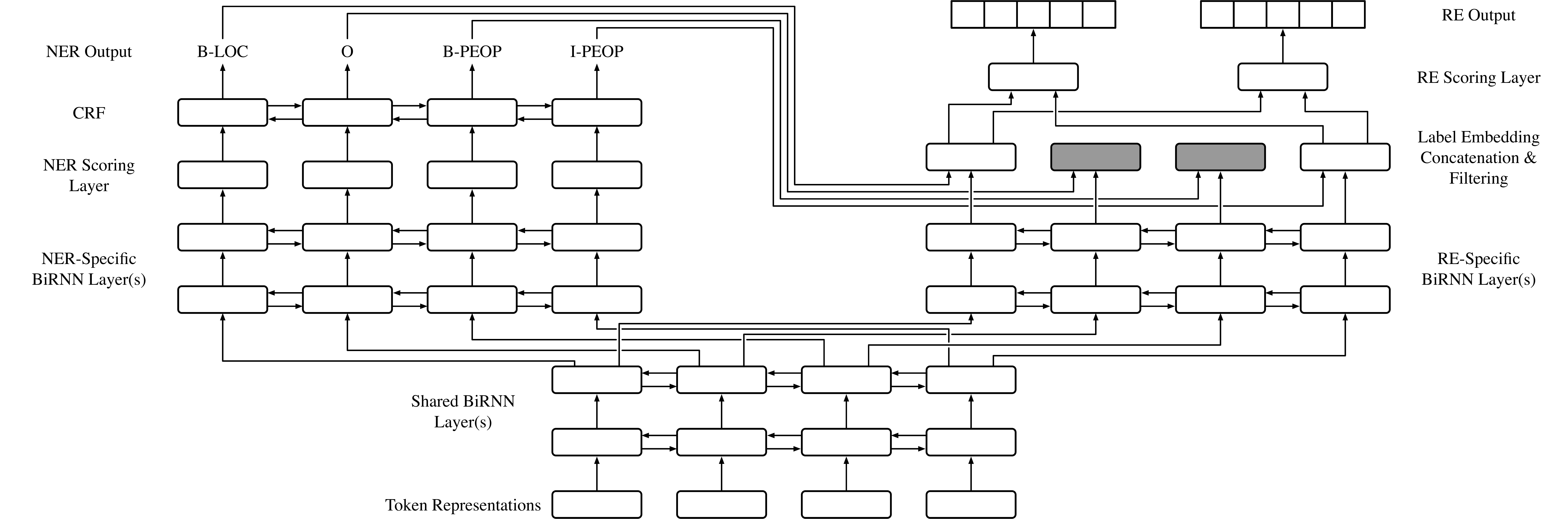}
    \caption{Illustration of our proposed architecture. Token representations are derived from a pre-trained ELMo model, pre-trained GloVe embeddings, learned character-based embeddings, and one-hot encoded casing vectors. The number of shared and task-specific BiRNN layers is treated as a hyperparameter of the model architecture. Only the final token in each entity span is used for predictions for the RE task; grey boxes indicate tokens that are not used for relation predictions. The output for the RE task is a vector of size \(|\mathcal{R}|\) for all pairs of entities, where \(\mathcal{R}\) is the set of all possible relations.}
    \label{fig:architecture}
\end{figure*}

A second method of incorporating greater task-specificity into these models is via deeper layers for solving the RE task. Miwa and Bansal~\shortcite{miwa-bansal-2016-end} and Li {\em et al.}~\shortcite{li2017neural} pass token representations through a BiLSTM layer and then use a softmax layer to label each token with the appropriate BILOU label. Both proposals then use a type of tree-structured bidirectional LSTM layer stacked on top of the shared BiLSTM to solve the RE task. Nguyen and Verspoor~\shortcite{nguyen2019end} use BiLSTM and CRF layers to perform the NER task. Label embeddings are created from predicted NER labels, concatenated with token representations, and then passed through a RE-specific BiLSTM. A biaffine attention layer \cite{dozat2016deep} operates on the output of this BiLSTM to predict relations.

An alternative to the BIO/BILOU scheme is the span-based approach, wherein spans of the input text are directly labeled as to whether they correspond to any entity and, if so, their entity types. Luan {\em et al.}~\shortcite{Luan_2018} adopt a span-based approach in which token representations are first passed through a BiLSTM layer. The output from the BiLSTM is used to construct representations of candidate entity spans, which are then scored for both the NER and RE tasks via feed forward layers. Luan {\em et al.}~\shortcite{Luan_2019} follow a similar approach, but construct coreference and relation graphs between entities to propagate information between entities connected in these graphs. The resulting entity representations are then classified for NER and RE via feed forward layers. To the best of our knowledge, the current SOTA model for joint NER and RE is the span-based proposal of Eberts and Ulges~\shortcite{eberts2019span}. In this architecture, token representations are obtained using a pre-trained BERT model that is fine-tuned during training. Representations for candidate entity spans are obtained by max pooling over all tokens in each span. Span representations are passed through an entity classification layer to solve the NER task. Representations of all pairs of spans that are predicted to be entities and representations of the contexts between these pairs are then passed through a final layer with sigmoid activation to predict relations between entities. With respect to their degrees of task-specificity, these span-based approaches resemble the BIO/BILOU approaches in which the majority of model parameters are shared, but each task possesses independent scoring and/or output layers. 

Overall, previous approaches to joint NER and RE have experimented little with deep task-specificity, with the exception of those models that include additional layers for the RE task. To our knowledge, no work has considered including additional NER-specific layers beyond scoring and/or output layers. This may reflect a residual influence of the pipeline approach in which the NER task must be solved first before additional layers are used to solve the RE task. However, there is no {\em a priori} reason to think that the RE task would benefit more from additional task-specific layers than the NER task. We also note that while previous work has tackled joint NER and RE in variety of textual domains, in all cases the number of shared and task-specific parameters is held constant across these domains.


\section{Model}

The architecture proposed here is inspired by several previous proposals \cite{katiyar-cardie-2017-going,bekoulis2018joint,nguyen2019end}. We treat the NER task as a sequence labeling problem using BIO labels. Token representations are first passed through a series of shared, BiRNN layers. Stacked on top of these shared BiRNN layers is a sequence of task-specific BiRNN layers for both the NER and RE tasks. We take the number of shared and task-specific layers to be a hyperparameter of the model. Both sets of task-specific BiRNN layers are followed by task-specific scoring and output layers. Figure \ref{fig:architecture} illustrates this architecture. Below, we use superscript \(e\) for NER-specific variables and layers and superscript \(r\) for RE-specific variables and layers.

\subsection{Shared Layers}

We obtain contextual token embeddings using the pre-trained ELMo 5.5B model \cite{peters2018deep}.\footnote{We also experimented with using a pre-trained BERT model rather than ELMo,  but performance when using ELMo was slightly higher than when using BERT. In order to help minimize the total number of trainable parameters in our model, we did not experiment with fine-tuning ELMo or BERT.} For each token in the input text \(t_i\), this model returns three vectors, which we combine via a weighted averaging layer. Each token \(t_i\)'s weighted ELMo embedding \(\mathbf{t}^{elmo}_{i}\) is concatenated to a pre-trained GloVe embedding \cite{pennington2014glove} \(\mathbf{t}^{glove}_{i}\), a character-level word embedding \(\mathbf{t}^{char}_i\) learned via a single BiRNN layer \cite{Lample_2016} and a one-hot encoded casing vector \(\mathbf{t}^{casing}_i\). The full representation of \(t_i\) is given by \(\mathbf{v}_i\) (where \(\circ\) denotes concatenation):
\begin{equation}
    \mathbf{v}_i = \mathbf{t}^{elmo}_{i} \circ \mathbf{t}^{glove}_{i} \circ \mathbf{t}^{char}_{i} \circ \mathbf{t}^{casing}_{i}
    \label{eq:weighted_avg}
\end{equation}
For an input text with \(n\) tokens, \(\mathbf{v}_{1:n}\) are fed as input to a sequence of one or more shared BiRNN layers, with the output sequence from the \(i\)th shared BiRNN layer serving as the input sequence to the \(i + 1\)st shared BiRNN layer. 

\subsection{NER-Specific Layers}

The final shared BiRNN layer is followed by a sequence of zero or more NER-specific BiRNN layers; the output of the final shared BiRNN layer serves as input to the first NER-specific BiRNN layer, if such a layer exists, and the output from from the \(i\)th NER-specific BiRNN layer serves as input to the \(i + 1\)st NER-specific BiRNN layer. For every token \(t_i\), let \(\mathbf{h}^{e}_i\) denote an NER-specific hidden representation for \(t_i\) corresponding to the \(i\)th element of the output sequence from the final NER-specific BiRNN layer or the final shared BiRNN layer if there are zero NER-specific BiRNN layers. An NER score for token \(t_i\), \(\mathbf{s}^{e}_i\), is obtained by passing \(\mathbf{h}^{e}_i\) through a series of two feed forward layers:
\begin{equation}
    \mathbf{s}^{e}_i = \text{FFNN}^{(e2)}\left((\text{FFNN}^{(e1)}(\mathbf{h}^{e}_i))\right)
    \label{eq:ner_score}
\end{equation}
The activation function of \(\text{FFNN}^{(e1)}\) and its output size are treated as hyperparameters. \(\text{FFNN}^{(e2)}\) uses linear activation and its output size is \(|\mathcal{E}|\), where \(\mathcal{E}\) is the set of possible entity types. The sequence of NER scores for all tokens, \(\mathbf{s}^{e}_{1:n}\), is then passed as input to a linear-chain CRF layer to produce the final BIO tag predictions, \(\hat{\mathbf{y}}^e_{1:n}\). During inference, Viterbi decoding is used to determine the most likely sequence \(\hat{\mathbf{y}}^e_{1:n}\).

\subsection{RE-Specific Layers}

Similar to the NER-specific layers, the output sequence from the final shared BiRNN layer is fed through zero or more RE-specific BiRNN layers. Let \(\mathbf{h}^{r}_i\) denote the \(i\)th output from the final RE-specific BiRNN layer or the final shared BiRNN layer if there are no RE-specific BiRNN layers. 

Following previous work \cite{gupta-etal-2016-table,bekoulis2018joint,nguyen2019end}, we predict relations between entities \(e_i\) and \(e_j\) using learned representations from the final tokens of the spans corresponding to \(e_i\) and \(e_j\). To this end, we filter the sequence \(\mathbf{h}^{r}_{1:n}\) to include only elements \(\mathbf{h}^{r}_{i}\) such that token \(t_i\) is the final token in an entity span. During training, ground truth entity spans are used for filtering. During inference, predicted entity spans derived from \(\hat{\mathbf{y}}^e_{1:n}\) are used. Each \(\mathbf{h}^{r}_{i}\) is concatenated to a learned NER label embedding for \(t_i\), \(\mathbf{l}^{e}_{i}\):
\begin{equation}
    \mathbf{g}^{r}_i = \textbf{h}^r_i \circ \mathbf{l}^e_i
    \label{eq:re_score_1}
\end{equation}
Ground truth NER labels are used to obtain \(\mathbf{l}^{e}_{1:n}\) during training, and predicted NER labels are used during inference.\footnote{Because ground truth NER labels are used to generate label embeddings during training, the output from the BiRNN layer(s) described in Section 3.2 is opaque to the RE-specific portion of the model during training. If predicted NER labels were used during training instead, as in \cite{nguyen2019end}, these layers would be shared rather than NER-specific.}

Next, RE scores are computed for every pair \((\mathbf{g}^{r}_i, \mathbf{g}^{r}_j)\). If \(\mathcal{R}\) is the set of possible relations, we calculate the \textsc{DistMult} score \cite{yang2014embedding} for every relation \(r_k \in \mathcal{R}\) and every pair \((\mathbf{g}^{r}_i, \mathbf{g}^{r}_j)\) as follows: 
\begin{equation}
    \textsc{DistMult}^{r_k}(\mathbf{g}^{r}_i, \mathbf{g}^{r}_j) = (\mathbf{g}^r_i)^T M^{r_k} \mathbf{g}^r_j
    \label{eq:distmult}
\end{equation}
\(M^{r_k}\) is a diagonal matrix such that \(M^{r_k} \in \mathbb{R}^{p \times p}\), where \(p\) is the dimensionality of \(\mathbf{g}^r_i\). We also pass each RE-specific hidden representation \(\mathbf{g}^{r}_i\) through a single feed forward layer: 
\begin{equation}
    \mathbf{f}^{r}_i = \text{FFNN}^{(r1)}(\mathbf{g}^{r}_i)
    \label{eq:re_score_1}
\end{equation}
As in the case of \(\text{FFNN}^{(e1)}\), the activation function of \(\text{FFNN}^{(r1)}\) and its output size are treated as hyperparameters. 

Let \(\textsc{DistMult}^r_{i,j}\) denote the concatenation of \(\textsc{DistMult}^{r_k}(\mathbf{g}^r_i, \mathbf{g}^r_j)\) for all \(r_k \in \mathcal{R}\) and let \(\cos_{i,j}\) denote the cosine distance between vectors \(\mathbf{f}^{r}_i\) and \(\mathbf{f}^{r}_j\). We obtain RE scores for \((t_i, t_j)\) via a feed forward layer:
\begin{equation}
    \mathbf{s}^r_{i,j} = \text{FFNN}^{(r2)}\left(\mathbf{f}^{r}_i \circ \mathbf{f}^{r}_j \circ \cos_{i,j} \circ \textsc{DistMult}^r_{i,j}\right)
\end{equation}
\(\text{FFNN}^{(r2)}\) uses linear activation, and its output size is \(|\mathcal{R}|\). Final relation predictions for a pair of tokens \((t_i, t_j)\), \(\hat{\mathbf{y}}^r_{i,j}\), are obtained by passing \(\mathbf{s}^r_{i,j}\) through an elementwise sigmoid layer. A relation is predicted for all outputs from this sigmoid layer exceeding \(\theta^r\), which we treat as a hyperparameter.

\subsection{Training}

During training, character embeddings, label embeddings, and weights for the weighted average layer, all BiRNN weights, all feed forward networks, and \(M^{r_k}\) for all \(r_k \in \mathcal{R}\) are trained in a supervised manner. As mentioned above, BIO tags for all tokens are used as labels for the NER task. For the the RE task, binary outputs are used. For every relation \(r_k \in R\) and for every pair of tokens \((t_i, t_j)\) such that \(t_i\) is the final token of entity \(e_i\) and \(t_j\) is the final token of entity \(e_j\), the RE label \(y^{r_k}_{i,j} = 1\) if \((e_i, e_j, r_k)\) is a true relation. Otherwise, we have \(y^{r_k}_{i,j} = 0\). 

For both output layers, we compute the cross-entropy loss. If \(\mathcal{L}_{NER}\) and \(\mathcal{L}_{RE}\) denote the cross-entropy loss for the NER and RE outputs, respectively, then the total model loss is given by \(\mathcal{L} = \mathcal{L}_{NER} + \lambda^r \mathcal{L}_{RE}\). The weight \(\lambda^r\) is treated as a hyperparameter and allows for tuning the relative importance of the NER and RE tasks during training. Final training for both datasets used a value of \(5\) for \(\lambda^r\). 

For the ADE dataset, we trained using the Adam optimizer with a mini-batch size of 16. For the CoNLL04 dataset, we used the Nesterov Adam optimizer with and a mini-batch size of \(2\). For both datasets, we used a learning rate of \(5\times10^{-4}\), During training, dropout was applied before each BiRNN layer, other than the character BiRNN layer, and before the RE scoring layer. 


\begin{table}[t]
\centering
\begin{tabular}{@{}lrr@{}}
\toprule
Hyperparameter & ADE & CoNLL04 \\
\midrule
BiRNN Type & GRU & GRU\\
Character BiRNN Size & 32 & 32\\
Non-Character BiRNN Size & 128 & 256 \\
\# Shared BiRNN Layers & 1 & 1 \\
\# NER-Specific BiRNN Layers & 1 & 1 \\
\# RE-Specific BiRNN Layers & 1 & 2 \\
\(\text{FFNN}^{(e1)}\) Activation & ReLU & tanh \\
\(\text{FFNN}^{(e1)}\) Output Size & 64 & 64 \\
\(\text{FFNN}^{(r1)}\) Activation & ReLU & ReLU \\
\(\text{FFNN}^{(r1)}\) Output Size & 128 & 128 \\
Label Embedding Size & 25 & 25 \\
Pre-BiRNN Dropout & 0.5 & 0.35 \\
Pre-RE Scoring Dropout & 0.5 & 0.5 \\
RE Threshold \(\theta^r\) & 0.9 & 0.9 \\
\bottomrule
\end{tabular}
\caption{Optimal hyperparameters used for final training on the ADE and CoNLL04 datasets.}
\label{tab:hpyerparameters}
\end{table}

\begin{table*}[t]
\centering
\begin{tabular}{@{}lllrrrrrrr@{}} \toprule
 Dataset & Model & Metric Type & \multicolumn{3}{c}{NER} & \multicolumn{3}{c}{RE} & Avg. \\ \cmidrule{4-10}
  &  & & Precision & Recall & F1 & Precision & Recall & F1 & F1 \\ \midrule
 ADE  & Bekoulis {\em et al.}~\shortcite{bekoulis2018joint} & Macro & 84.72 & 88.16 & 86.40 & 72.10 & 77.24 & 74.58 & 80.49 \\
      & Eberts and Ulges~\shortcite{eberts2019span} & Macro & \textbf{89.26} & 89.26 & 89.25 & 78.09 & 80.43 & 79.24 & 84.25 \\
      & \textbf{Ours} & Macro & 89.06 & \textbf{89.63} & \textbf{89.48} & \textbf{80.51} & \textbf{86.81} & \textbf{83.74} & \textbf{86.61} \\
\midrule
CoNLL04   & Nguyen and Verspoor~\shortcite{nguyen2019end}  & Macro & -- & -- & 86.20 & -- & -- & 64.40 & 75.30 \\
& Li {\em et al.}~\shortcite{li2019entity}  & Micro & 89.00 & 86.60 & 87.80 & 69.20 & 68.20 & 68.90 & 78.35 \\
& Eberts and Ulges~\shortcite{eberts2019span} & Macro & 85.78 & \textbf{86.84} & 86.25 & 74.75 & \textbf{71.52} & \textbf{72.87} & 79.56\\
& Eberts and Ulges~\shortcite{eberts2019span} & Micro & 88.25 & 89.64 & 88.94 & 73.04 & \textbf{70.00} & \textbf{71.47} & 80.21\\
& \textbf{Ours} & Macro & \textbf{87.92} & 86.42 & \textbf{87.00} & \textbf{77.73} & 68.38 & 72.63 & \textbf{79.82} \\
& \textbf{Ours} & Micro & \textbf{89.84} & \textbf{89.73} & \textbf{89.78} & \textbf{78.69} & 64.84 & 71.08 & \textbf{80.43} \\
\bottomrule
 \end{tabular}
\caption{Precision, Recall, and F1 scores for our model and other recent models on the ADE and CoNLL04 datasets. Because our scores are averaged across multiple trials, F1 scores shown here cannot be directly calculated from the precision and recall scores shown here. Note that Nguyen and Verspoor do not report precision and recall scores.}
\label{tab:results}
\end{table*}

\section{Experiments}

We evaluate the architecture described above using the following two publicly available datasets.

\subsubsection{ADE}
The Adverse Drug Events (ADE) dataset \cite{gurulingappa2012development} consists of 4,272 sentences describing adverse effects from the use of particular drugs. The text is annotated using two entity types ({\em Adverse-Effect} and {\em Drug}) and a single relation type ({\em Adverse-Effect}). Of the entity instances in the dataset, 120 overlap with other entities. Similar to prior work using BIO/BILOU tagging, we remove overlapping entities. We preserve the entity with the longer span and remove any relations involving a removed entity. 

There are no official training, dev, and test splits for the ADE dataset, leading previous researchers to use some form of cross-validation when evaluating their models on this dataset. We split out 10\% of the data to use as a held-out dev set. Final results are obtained via 10-fold cross-validation using the remaining 90\% of the data and the hyperparameters obtained from tuning on the dev set. Following previous work, we report macro-averaged performance metrics averaged across each of the 10 folds.

\subsubsection{CoNLL04}

The CoNLL04 dataset \cite{roth2004linear} consists of 1,441 sentences from news articles annotated with four entity types ({\em Location}, {\em Organization}, {\em People}, and {\em Other}) and five relation types ({\em Works-For}, {\em Kill}, {\em Organization-Based-In}, {\em Lives-In}, and {\em Located-In}). This dataset contains no overlapping entities.

We use the three-way split of \cite{gupta-etal-2016-table}, which contains 910 training, 243 dev, and 288 test sentences. All hyperparameters are tuned against the dev set. Final results are obtained by averaging results from five trials with random weight initializations in which we trained on the combined training and dev sets and evaluated on the test set. As previous work using the CoNLL04 dataset has reported both micro- and macro-averages, we report both sets of metrics.

\vspace{7pt}

In evaluating NER performance on these datasets, a predicted entity is only considered a true positive if both the entity's span and span type are correctly predicted. In evaluating RE performance, we follow previous work in adopting a {\em strict} evaluation method wherein a predicted relation is only considered correct if the spans corresponding to the two arguments of this relation and the entity types of these spans are also predicted correctly. We experimented with LSTMs and GRUs for all BiRNN layers in the model and experimented with using \(1-3\) shared BiRNN layers and \(0-3\) task-specific BiRNN layers for each task. Hyperparameters used for final training are listed in Table \ref{tab:hpyerparameters}.


\subsection{Results}

Full results for the performance of our model, as well as other recent work, are shown in Table \ref{tab:results}. In addition to precision, recall, and F1 scores for both tasks, we show the average of the F1 scores across both tasks. On the ADE dataset, we achieve SOTA results for both the NER and RE tasks. On the CoNLL04 dataset, we achieve SOTA results on the NER task, while our performance on the RE task is competitive with other recent models. On both datasets, we achieve SOTA results when considering the average F1 score across both tasks. The largest gain relative to the previous SOTA performance is on the RE task of the ADE dataset, where we see an absolute improvement of 4.5 on the macro-average F1 score.

While the model of Eberts and Ulges~\shortcite{eberts2019span} outperforms our proposed architecture on the CoNLL04 RE task, their results come at the cost of greater model complexity. As mentioned above, Eberts and Ulges fine-tune the BERT\textsubscript{BASE} model, which has 110 million trainable parameters. In contrast, given the hyperparameters used for final training on the CoNLL04 dataset, our proposed architecture has approximately 6 million trainable parameters.

The fact that the optimal number of task-specific layers differed between the two datasets demonstrates the value of taking the number of shared and task-specific layers to be a hyperparameter of our model architecture. As shown in Table \ref{tab:hpyerparameters}, the final hyperparameters used for the CoNLL04 dataset included an additional RE-specific BiRNN layer than did the final hyperparameters used for the ADE dataset. We suspect that this is due to the limited number of relations and entities in the ADE dataset. For most examples in this dataset, it is sufficient to correctly identify a single {\em Drug} entity, a single {\em Adverse-Effect} entity, and an {\em Adverse-Effect} relation between the two entities. Thus, the NER and RE tasks for this dataset are more closely related than they are in the case of the CoNLL04 dataset. Intuitively, cases in which the NER and RE problems can be solved by relying on more shared information should require fewer task-specific layers. 

\subsection{Ablation Study}

To further demonstrate the effectiveness of the additional task-specific BiRNN layers in our architecture, we conducted an ablation study using the CoNLL04 dataset. We trained and evaluated in the same manner described above, using the same hyperparameters, with the following exceptions:
\begin{enumerate}
    \item We used either (i) zero NER-specific BiRNN layers, (ii) zero RE-specific BiRNN layers, or (iii) zero task-specific BiRNN layers of any kind.
    \item We increased the number of shared BiRNN layers to keep the total number of model parameters consistent with the number of parameters in the baseline model.
    \item We average the results for each set of hyperparameter across three trials with random weight initializations. 
\end{enumerate}

Table \ref{tab:ablation} contains the results from the ablation study. These results show that the proposed architecture benefits from the inclusion of both NER- and RE-specific layers. However, the RE task benefits much more from the inclusion of these task-specific layers than does the NER task. We take this to reflect the fact that the RE task is more difficult than the NER task for the CoNLL04 dataset, and therefore benefits the most from its own task-specific layers.  This is consistent with the fact that the hyperparameter setting that performs best on the RE task is that with no NER-specific BiRNN layers, i.e. the setting that retained RE-specific BiRNN layers. In contrast, the inclusion of task-specific BiRNN layers of any kind had relatively little impact on the performance on the NER task. 

Note that the setting with no NER-specific layers is somewhat similar to the setup of Nguyen and Verspoor's~\shortcite{nguyen2019end} model, but includes an additional shared and an additional RE-specific layer. That this setting outperforms Nguyen {\em et al.}'s model reflects the contribution of having deeper shared and RE-specific layers, separate from the contribution of NER-specific layers.

\begin{table}[t]
\centering
\begin{tabular}{@{}llrr@{}}
\toprule
Model Type & Metric Type & NER F1 & RE F1\\ \midrule
\textbf{Baseline} & Macro & \textbf{87.00} & \textbf{72.63} \\ 
 & Micro & \textbf{89.78} & \textbf{71.08} \\ \midrule
No NER-Specific  & Macro & 86.19 & 68.45 \\
BiRRNs & Micro & 89.24 & 67.05 \\
No RE-Specific & Macro & 85.84 & 59.57  \\
BiRRNs & Micro & 89.06 & 57.82 \\
No Task-Specific & Macro & 86.16 & 57.45 \\
BiRRNs & Micro & 89.46 & 55.47 \\
\bottomrule
\end{tabular}
\caption{Results from an ablation study using the CoNLL04 dataset. All models have the same number of total parameters.}
\label{tab:ablation}
\end{table}

\section{Conclusion}

Our results demonstrate the utility of using deeper task-specificity in models for joint NER and RE and of tuning the level of task-specificity separately for different datasets. We conclude that prior work on joint NER and RE undervalues the importance of task-specificity. More generally, these results underscore the importance of correctly balancing the number of shared and task-specific parameters in MTL.

We note that other approaches that employ a single model architecture across different datasets are laudable insofar as we should prefer models that can generalize well across domains with little domain-specific hyperparameter tuning. On the other hand, the similarity between the NER and RE tasks varies across domains, and improved performance can be achieved on these tasks by tuning the number of shared and task-specific parameters. In our work, we treated the number of shared and task-specific layers as a hyperparameter to be tuned for each dataset, but future work may explore ways to select this aspect of the model architecture in a more principled way. For example, Vandenhende {\em et al.}~\shortcite{vandenhende2019branched} propose using a measure of affinity between tasks to determine how many layers to share in MTL networks. Task affinity scores of NER and RE could be computed for different textual domains or datasets, which could then guide the decision regarding the number of shared and task-specific layers to employ for joint NER and RE models deployed on these domains.  

Other extensions to the present work could include fine-tuning the model used to obtain contextual word embeddings, e.g. ELMo or BERT, during training. In order to minimize the number of trainable parameters, we did not employ such fine-tuning in our model, but we suspect a fine-tuning approach could lead to improved performance relative to our results. An additional opportunity for future work would be an extension of this work to other related NLP tasks, such as co-reference resolution and cross-sentential relation extraction.

\bibliographystyle{named}
\bibliography{deeper_task_specificity}

\end{document}